\documentclass[letterpaper, 10 pt, conference]{ieeeconf}  

\usepackage{amsfonts}
\usepackage{algorithm}
\usepackage{algpseudocode}
\usepackage{graphicx}
\usepackage{hyperref}
\usepackage{subcaption}

\IEEEoverridecommandlockouts                              

\title{\LARGE \bf
Occlusion-Aware 2D and 3D Centerline Detection for Urban Driving via Automatic Label Generation
}

\author{David Paz$^{*\dagger}$, Narayanan E. Ranganatha$^{*\dagger}$, Srinidhi K. Srinivas$^{\dagger}$, Yunchao Yao$^{\dagger}$ and Henrik I. Christensen$^{\dagger}$
\thanks{*Denotes equal contribution.}
\thanks{$^{\dagger}$Contextual Robotics Institute, University of California San Diego, La
Jolla, CA 92093, USA}
}

\begin{document}

\maketitle
\thispagestyle{empty}
\pagestyle{empty}

\begin{abstract}

This research work seeks to explore and identify strategies that can determine road topology information in 2D and 3D under highly dynamic urban driving scenarios. To facilitate this exploration, we introduce a substantial dataset comprising nearly one million automatically labeled data frames. A key contribution of our research lies in developing an automatic label-generation process and an occlusion handling strategy. This strategy is designed to model a wide range of occlusion scenarios, from mild disruptions to severe blockages. Furthermore, we present a comprehensive ablation study wherein multiple centerline detection methods are developed and evaluated. This analysis not only benchmarks the performance of various approaches but also provides valuable insights into the interpretability of these methods. Finally, we demonstrate the practicality of our methods and assess their adaptability across different sensor configurations, highlighting their versatility and relevance in real-world scenarios. Our dataset and experimental models are publicly available. \footnote{\href{https://github.com/AutonomousVehicleLaboratory/cl_det}{https://github.com/AutonomousVehicleLaboratory/cl\_det}}

\end{abstract}

\section{INTRODUCTION}

The nature of the world that autonomous vehicles operate in is highly dynamic. To be able to successfully and safely navigate in these dynamic environments, several challenges need to be addressed, such as detection \cite{li2022bevformer, yang2022deepinteraction}, tracking \cite{camo-mot, msmd}, prediction \cite{shi2022mtr, multipath}, and planning \cite{casas2021mp3, mats} to name a few. 

Recently, contributions to the field of autonomous vehicle technology from the open-source community have played a significant role in addressing some of these challenges. These contributions have come in the form of open-source datasets, such as Argoverse2~\cite{Argoverse2}, Waymo Open Dataset~\cite{sun2020scalability}, and NuPlan~\cite{nuplan}. These datasets contain meticulously labeled sensor data and high-definition mapping information, among a multitude of other information required by an autonomous vehicle to safely and reliably navigate.

High-definition maps are a crucial component within these datasets, and they are integral to the operation of autonomy stacks, as demonstrated by methods like \cite{autoware, apollo17, micro-mobility}, which facilitate global planning for point-to-point navigation. These maps provide detailed lane-level definitions and road network connectivity information, offering context for prediction models and aiding planning models in trajectory generation and optimization tasks.

However, it's important to note that these applications often operate under the assumption of a static world. In practice, this assumption can be challenged by changes in the road layout or temporary constructions, leading to potential failures. Given the substantial disparities between these static map definitions and the evolving real-world conditions, autonomous agents would struggle to devise feasible trajectories without continuously updating their knowledge of the static map features. This underscores the critical importance of dynamic and real-time scene comprehension.

\begin{figure}[t]
  \centering
   \includegraphics[width=\linewidth]{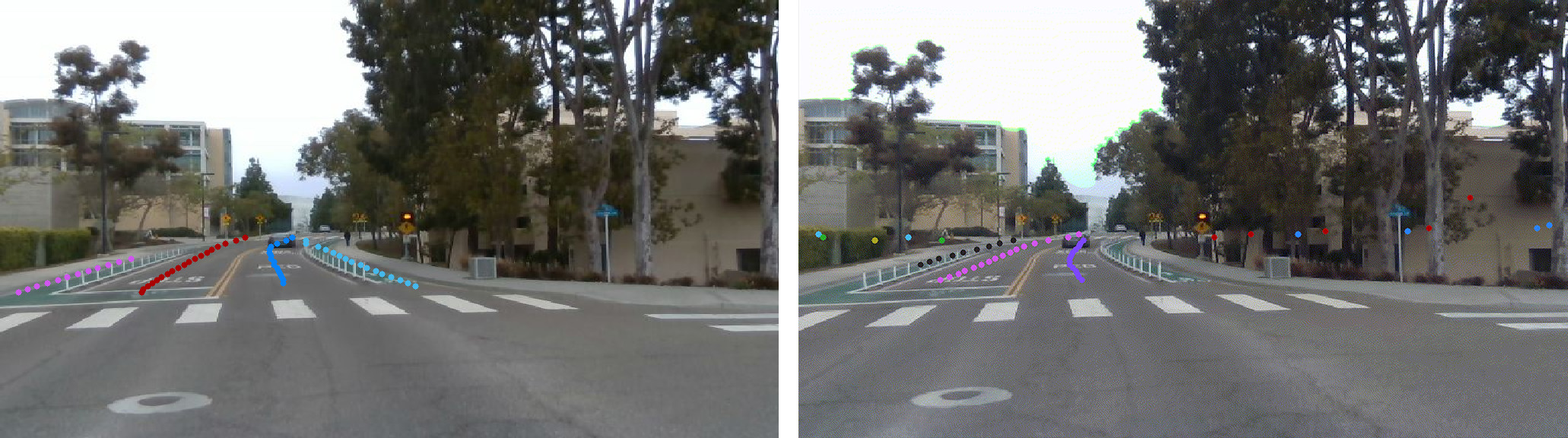}

   \caption{Centerline prediction outputs for urban driving scenario with (left) and without (right) occlusion handling logic. The model without occlusion handling logic hallucinates in regions without centerline features.}
   \label{fig:ucsd-viz}
\end{figure}

To explore dynamic scene modeling strategies, this work presents three key contributions to the open-source research community, namely
\begin{itemize}
    \item A publicly available dataset automatically labeled by utilizing vector map data from the Argoverse2 sensor dataset. We present an automatic image labeling pipeline that leverages  ego-vehicle localization, the state information of each camera, projective geometry to align map features with image data, and semantic segmentation to decorate 2D and 3D keypoint correspondences with semantic class attributes. The dataset contains over 900,000 annotated image frames across three front-facing cameras.
    \item Using contextual information and per-keypoint semantics, we propose an occlusion handling formulation to mitigate the artifacts and hallucinations generated from dynamic objects or large static structures occluding the camera view.
    \item We conduct extensive ablation studies on the Argoverse2 dataset to evaluate the performance of multiple centerline detection models under various occlusion resolution constraints, while highlighting key aspects of interpretability. Finally, a qualitative analysis is performed on the models’ effectiveness in unseen sensor configurations and scenes, i.e. Figure~\ref{fig:ucsd-viz}.
\end{itemize}

\section{Related Work}
Lane detection and semantic segmentation have been active areas of research for decades. Early investigations~\cite{dickmanns07} relied on conventional edge detection techniques for applications in autonomous driving. In recent times, there has been a growing inclination towards using learning-based methods in the domain of lane detection and segmentation. These methods encompass semantic segmentation and incorporate approaches based on keypoint detection and parameterized representations. Semantic segmentation, as a task, involves the classification of individual pixels in an image into predefined classes. Publicly accessible datasets like Mapillary~\cite{MVD2017_Mapillary_Vistas} and Cityscapes~\cite{Cordts_2016_CVPR_cityscapes} have played a role in fostering a multitude of open-source contributions within the domain of semantic segmentation, specifically for applications in autonomous driving. These initiatives leverage various tools, including encoder-decoder frameworks~\cite{unet}, spatial pyramid pooling modules~\cite{deeplabv3},  and attention mechanisms directed at resolving intricate details~\cite{deeplabv3}. Despite their capabilities, it is important to note that semantic segmentation can impose a substantial computational burden both during training and inference, potentially limiting its applicability in real-time scene comprehension.

An alternative strategy for inferring drivable regions and trajectories centers around the prediction of keypoints and splines to delineate lanes directly. PINet~\cite{ko2021key} exemplifies real-time image-based keypoint detection, while parameterized spline representations like Bézier curves~\cite{feng2022rethinking} have also found applications in lane detection. These representations aim to compress lane information using a fixed number of control points; however, they may fall short when dealing with highly irregular lanes or sharp curves. More recently, the scope of lane detection has expanded into the realm of 3D~\cite{curveformer} through the utilization of synthetic~\cite{guo2020gen} and real-world datasets~\cite{chen2022persformer}.

In contrast to prior research, our approach primarily emphasizes directly predicting centerline keypoints, as demonstrated in Figure \ref{fig:ucsd-viz}. This approach yields significant advantages for downstream applications and post-processing. In practical terms, the centerlines of each lane serve as reference trajectories that motion planners can directly harness. Conventional lane detection and segmentation methods still necessitate additional steps for centerline extraction and reasoning, depending on the scene context and the traversability of bounded lane markings. A recent development in centerline detection has been explored in the context of HD map generation~\cite{maptrv2, li2023graphbased, wang2023openlanev2}. This direction uses surround view cameras to estimate the bird's-eye-view road topology in an end-to-end fashion. However, our approach for centerline detection differs in two aspects: i) we focus on centerline detection in 2D and 3D from camera-centric views to enable late-fusion methods across multi-view camera perspectives, ii) the approach employed explicitly handles occlusion from dynamic objects and large structures to reduce hallucinations or improbable outputs. The design of our strategy presents high potential in the next generation of scene understanding and perception frameworks based on validations from real-world data and unseen environments.

\section{Methods}

Our main objective is to build a system that can process an image $\mathbf{I} \in \mathbb{R}^{H\times W\times 3}$ and predict 2D and 3D centerlines, which can then be used for real-time navigation tasks. It is important to emphasize that 
the system must rely on fewer priors during inference to make it suitable for operation in dynamic urban driving tasks--which may be subject to continuous changes and lane-level updates. Our approach comprises three steps: $\mathbf{A)}$ Automatic centerline label generation using map data; $\mathbf{B)}$ Semantic segmentation-based occlusion handling to reason about light to severe occlusion scenes; $\mathbf{C)}$ Model development focusing on real-time execution. 

\subsection{Automatic centerline label generation} \label{sec:dataset}

We augment the Argoverse2 dataset to create a dataset containing 942,161 individually labeled frames across the three front-facing cameras with known camera parameters. In contrast to standard labeling techniques that generally involve extensive manual labeling, we designed a pipeline that automatically generates image keypoint features with ground truth depth information. First, the ego-vehicle pose is synchronized to each camera by utilizing LiDAR data; given the pose of each camera over time and performing pose corrections, we align vector map information defined in a city coordinate frame and project its 3D features into each image frame. Given that projection preserves 2D and 3D correspondences, this facilitates assigning unique lane IDs. 

The dataset includes labels for centerlines and lane boundaries; where each label is interpolated in 3D and post-processed in an image frame to remove features that are too far or too tightly packed after projection. An important consideration involves intersections: although the full road network connectivity for each intersection is available, we find that sufficient context to identify entry and exit points of each intersection lane is often lacking when the vehicle is at an intersection. For this reason, we additionally remove labels that are too short or are part of intersections. The total split for the train, validation, and test sets corresponds to 659,720, 141,138, and 141,303 image frames, respectively.

\subsection{Semantic Segmentation-based Occlusion Handling}

\begin{figure}[t]
  \vspace{0.2cm}
  \centering
   \includegraphics[width=.7\linewidth]{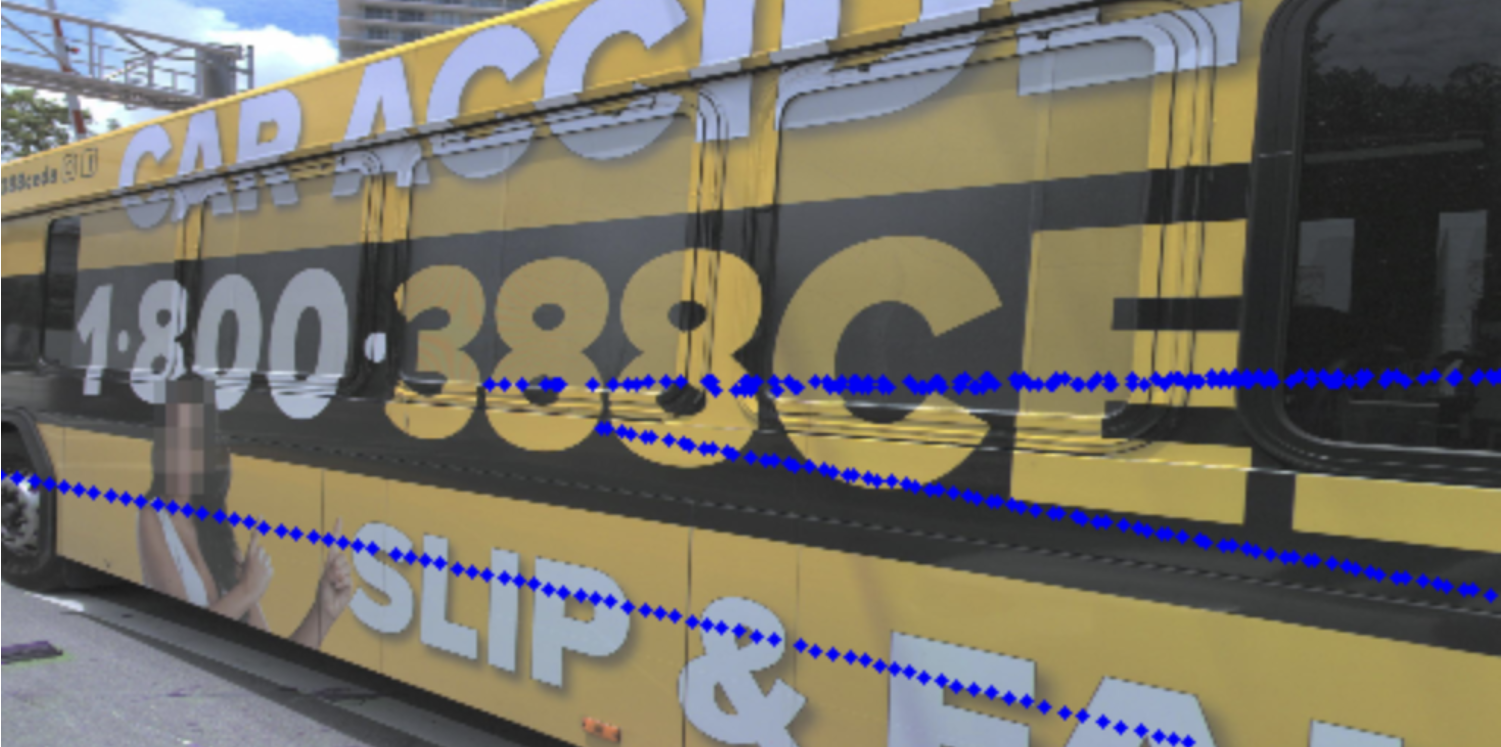}

   \caption{Data sample generated by the automatic labelling process without modeling occlusion.}
   \label{fig:occ-eg-1}
\end{figure}

\begin{figure}[t]
  \centering
   \includegraphics[width=\linewidth]{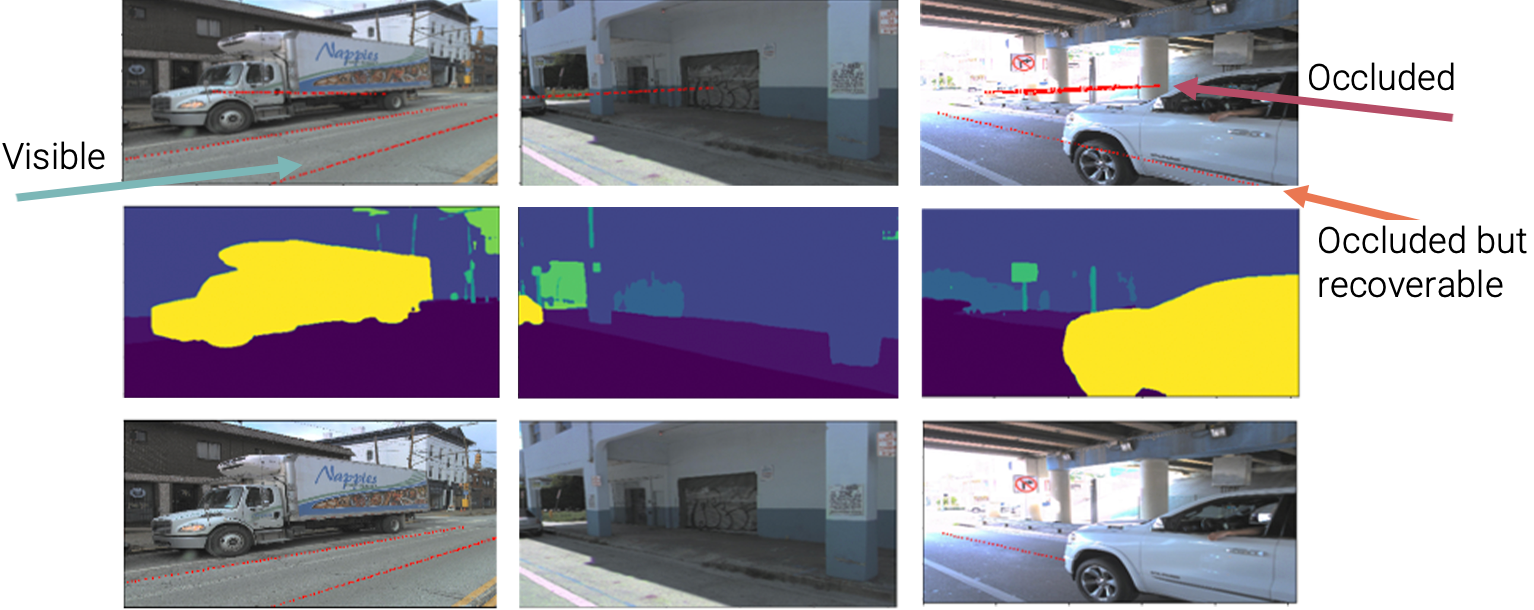}

   \caption{Occlusion handling for centerlines with and without sufficient context. Original centerline projections are shown in the first row, semantic segmentation outputs are shown in the second row, and centerlines with an occlusion free logic are shown in the third row. Arrows denote three different types of occlusion modes.}
   \label{fig:occ-removal}
\end{figure}

One challenge that is observed in training models using the data generated via the above process involves occlusion. Since the map information is directly utilized to automatically label images based on projective geometry, some road structures and moving agents can severely occlude the centerline features of interest. An example can be visualized in Figure \ref{fig:occ-eg-1}: while the centerline features from the nearby road elements have valid projections regarding the camera’s field of view, the keypoints are fully occluded by a bus and cannot be realistically recovered. However, some centerlines may only be partially occluded, and the image may provide sufficient context to recover such lines such as the one presented in Figure \ref{fig:occ-removal}. Figure \ref{fig:occ-removal} shows the three occlusion modes for centerlines that are observed in the dataset and that can be differentiated via occlusion handling.







Hence, our approach to occlusion handling is motivated by a combination of occlusion severity and context. Occlusion by objects within context can allow us to recover centerline estimates even in the presence of severe occlusion. On the other hand, objects that provide little context, such as buildings, can be difficult to recover as shown in Fig.~\ref{fig:occ-eg-1}. The observations motivate our design that leverages state-of-the-art semantic segmentation methods to quantify occlusion using different types of object classes. The semantic segmentation process leverages the model introduced in~\cite{tao2020hierarchical} which is trained on the Mapillary and Cityscapes datasets. The model processes individual image frames and decorates the 2D and 3D centerline features based on the semantic segmentation masks generated and their overlap with the keypoints. A benefit of the approach lies in the performance and robustness of the hierarchical multiscale attention method of the network. This attention approach achieves state-of-the-art performance by combining the outputs of a single input processed at different scales (i.e. 0.5x, 1.0x, and 2.0x).

\begin{figure*}[t]
  \vspace{0.2cm}
  \centering
   \includegraphics[width=.8\textwidth]{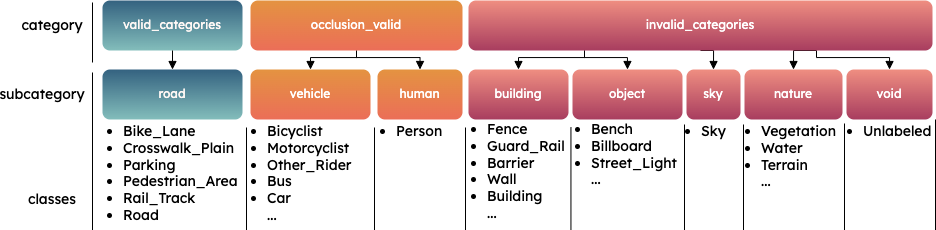}

   \caption{Category ontology for occlusion handling}
   \label{fig:catont}
\end{figure*}

We characterize occlusions from objects based on three different categories. The first one corresponds to occlusion from objects with insufficient context, i.e., objects away from the road that cannot be drawn for centerline prediction (invalid). The second one captures occlusion from on-road objects such as pedestrians and vehicles (occlusion valid) that are directly related to the driving task and contain contextual information that can help recover centerline features. Finally, the third one corresponds to road-level semantic classes, such as road and crosswalk (valid). For simplicity, we refer to these categories as invalid, occlusion\_valid, and valid and utilize the well-established Mapillary dataset semantic classes in the formulation of these categories. The Mapillary dataset contains up to 124 semantic object classes. From these classes, we cluster individual classes into nine different subcategories based on their semantic meanings and assign each subcategory into the valid, occlusion valid, or invalid parent categories as shown in Figure \ref{fig:catont}. This provides a straightforward formulation for calculating percent occlusion, as outlined in Algorithm \ref{alg:filter_keypoints}.

To model these occlusions, we calculate the occlusion percentage $R_{occ}$ for each centerline. We first calculate the number of pixels that are occluded $N_{occ}$. Let $P_{init} = \{p_1, p_2 \dots p_N\}$ be the keypoints belonging to the centerline before occlusion handling, and $P_{final}$ be the keypoints belonging to it after occlusion handling. $N$ is the number of keypoints in the centerline. If a keypoint (in pixel representation) $p_i$ belongs to the valid class, then $p_i \in P_{final}$. Any pixel $p_i$ is considered occluded if it belongs to the occlusion\_valid or invalid classes. If $p_i$ is part of a occlusion\_valid class, then $p_i \in P_{final}$, but we also increment $N_{occ}$. If $p_i$ is part of an invalid class, then $p_i \notin P_{final}$ and we also increment $N_{occ}$. Finally, we calculate the occlusion percentage $R_{occ}$ as:
\begin{equation}
    R_{occ} = \frac{N_{occ}}{N}
\end{equation}

We also define an occlusion threshold $T_{occ}$ as a hyperparameter and if $R_{occ} \ge T_{occ}$, the centerline is removed from the training set. It is also important to note that this is used only during training to filter the ground truth and does not play a role at inference time.

\begin{algorithm}
\caption{Filtering Keypoints Algorithm}\label{alg:filter_keypoints}
\begin{algorithmic}[1]
    \State \textbf{Input:} $P$, $P_{labels}$, \textit{invalid}, \textit{occlusion\_valid}, \textit{valid} labels, semantic labels \textit{labels}, Occlusion threshold $T_{occ}$,
    \State \textbf{Output:} Filtered centerlines $P_{filtered}$
    \State Initialize empty list: $P_{filtered}$
    \For{$i$ \textbf{in range} \textit{len($P$)}}\hfill \# loop over centerlines 
        \State $P_{init} \gets P[i]$ \hfill \# centerline keypoints
        \State $p\_labels \gets P_{labels}[i]$ \hfill \# keypoint semantic labels
        \State $N \gets \textit{len}(P_{init})$
        \State $N_{occ} \gets 0$
        \State $P_{final} \gets []$
        \For{$j$ \textbf{in range} \textit{len($P_{init}$)}}
            \State $label \gets \textit{labels}[p\_labels[j]]$
            \If{$\textit{label.category} \in \textit{invalid}$}
                \State $N_{occ} \gets N_{occ} + 1$
                \State \textbf{continue}
            \ElsIf{$\textit{label.category} \in \textit{occlusion\_valid}$}
                \State $N_{occ} \gets N_{occ} + 1$
                \State $P_{final}.append(P_{init}[j])$
            \ElsIf{$\textit{label.category} \in \textit{valid}$}
                \State $P_{final}.append(P_{init}[j])$
            \EndIf
        \EndFor
        \State $R_{occ} = \frac{N_{occ}}{N}$
        \If{$R_{occ} < T_{occ}$}
            \State $P_{filtered}.append(P_{final})$
        \EndIf
    \EndFor
\end{algorithmic}
\end{algorithm}

\subsection{2D-3D Centerline Detection Baseline}

\begin{figure*}[t]
  \centering
   \includegraphics[width=.9\textwidth]{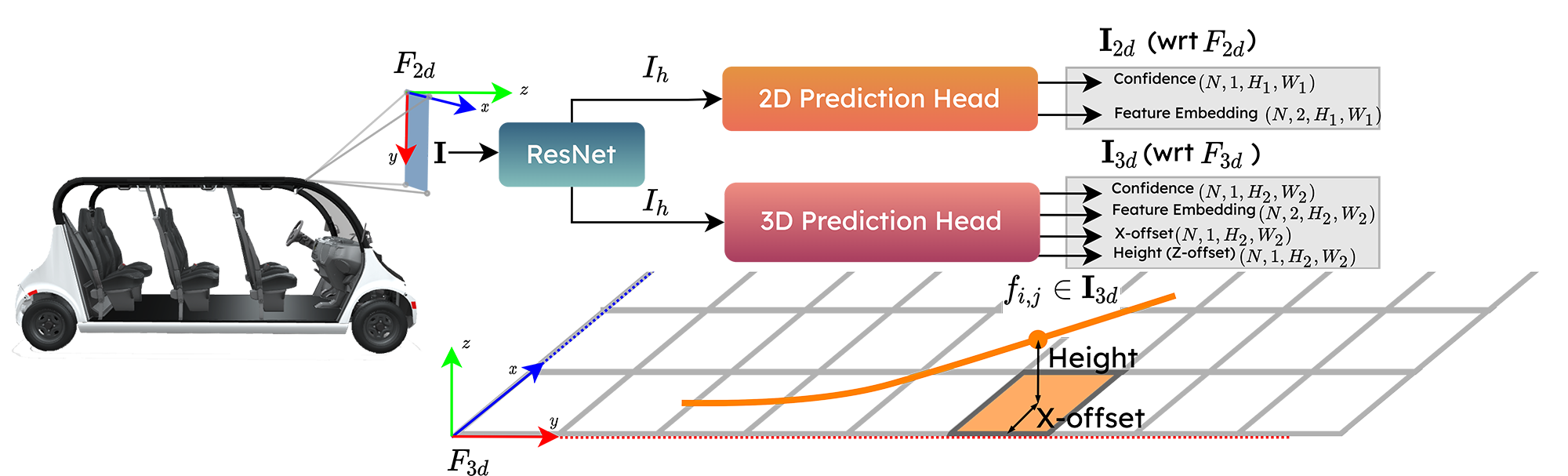}
   \caption{Reference frames and architecture details used for 2D and 3D centerline estimation.}
   \label{fig:architecture}
\end{figure*}

Our approach shown in Figure~\ref{fig:architecture} is motivated by the recently released BEV-LaneDet \cite{wang2022bev}. We use a feature extractor to generate the features of the front-view image. As in BEV-LaneDet, we use ResNet34 as the feature extractor. The output of the feature extractor is fed into 2D and 3D prediction heads. The 2D prediction head is used as auxiliary supervision and outputs a confidence map and feature embedding. The 3D prediction head outputs a confidence map, feature embedding, X-offset, and the Z-offset(Height). These are described in detail below.

\subsubsection{2D Prediction Head}

The 2D Prediction Head is based on the architecture used for LaneNet \cite{LaneNet}. It consists of the segmentation branch and the embedding branch. The segmentation branch is used to produce a binary lane mask. The embedding branch is used to produce a N-dimensional embedding per pixel. The embedding is trained via distance metric learning such that pixels that belong to the same lane have smaller distances in the embedding space, while the pixels belonging to different lanes have larger distances in the embedding space. This is done via the push and pull losses introduced in \cite{LaneNet} shown in Equations \ref{eq:2dpull}-\ref{eq:2dembed}.
\begin{equation}\label{eq:2dpull}
    \mathcal{L}_{pull} = \frac{1}{C}\sum_{c=1}^{C} \frac{1}{N_C}\sum_{i=1}^{N_C}\left[||\mu_c - x_i|| - \delta_{pull}\right]^2_{+}
\end{equation}
\begin{equation}\label{eq:2dpush}
    \mathcal{L}_{push} = \frac{1}{C(C-1)}\sum_{c_A=1}^{C}\sum_{c_B=1,c_B \ne c_A}^{C} \left[ \delta_{push} - ||\mu_{c_A} - \mu_{c_B}||\right]^2_{+}
\end{equation}
\begin{equation}\label{eq:2dembed}
    \mathcal{L}_{2D}^{embed} = \lambda_{2D}^{pull}\mathcal{L}_{pull} + \lambda_{2D}^{push}\mathcal{L}_{push}
\end{equation}

with $[x]_+$ = max(0,x)$. \delta_{pull}$ and $\delta_{push}$ are hyperparameters. If the embedding distance between the cluster center $\mu_c$ and the pixel embedding is $<\delta_{pull}$ then $\mathcal{L}_{pull} = 0$ and if the embedding distance between the cluster center and the pixel embedding is $>\delta_{push}$ then $\mathcal{L}_{push} = 0$. $C$ denotes the total number of classes.

The segmentation branch is trained with the binary cross-entropy loss function with inverse class weighting due to the high imbalance between the centerline and background classes. We refer to this loss as $\mathcal{L}_{2D}^{seg}$. Therefore, we have:
\begin{equation}
    \mathcal{L}_{2D} = \mathcal{L}_{2D}^{embed} + \lambda_{2D}^{seg}\mathcal{L}_{2D}^{seg}
\end{equation}

\subsubsection{3D Prediction Head}

The 3D Prediction Head first converts the features from the feature extractor into Birds-Eye-View(BEV). We conduct experiments using two modules to achieve this conversion. First is the View Relation Module (VRM) introduced in~\cite{wang2022bev} and the other is attention-based approach using the Transformer Encoder as used in~\cite{50650}. Once this conversion from image frame to BEV frame is done, these features are used by four branches, the segmentation branch, the X-offset branch, the Z-offset branch, and the embedding branch. The BEV plane ($z = 0$) is divided into a grid of cells (the size of each cell is defined as $0.5\times0.5m^2$ as recommended in~\cite{wang2022bev}). Let $s_1\times s_2$ be the size of this grid. The segmentation branch is similar to the segmentation branch of the 2D Prediction Head but it operates on the BEV grid. It is trained with the binary cross-entropy loss function and is referred to as $\mathcal{L}_{seg}^{3D}$. The X-offset branch predicts the offset from the center of the grid cell for the cells through which the lane passes. This branch is trained via a standard MSE loss to achieve finer-grained precision and is referred to as $\mathcal{L}_{offset}^{3D}$. The Z-offset branch predicts the height of the lane as this cannot be predicted by the previously mentioned branches. This branch is also trained via a standard MSE loss and is referred to as $\mathcal{L}_{height}^{3D}$. The embedding branch is trained very similarly to the 2D prediction head embedding via distance metric learning such that grid cells that belong to the same lane have smaller distances in the embedding space while the grid cells belonging to different lanes have larger distances in the embedding space. The losses are in Equations \ref{eq:3dpull}-\ref{eq:3dheight}.

\begin{equation}\label{eq:3dpull}
    \mathcal{L}_{pull}^{3D} = \frac{1}{C}\sum_{c=1}^{C} \frac{1}{N_C}\sum_{i=1}^{N_C}\left[||\mu_c - x_i|| - \delta_{pull}\right]^2_{+}
\end{equation}
\begin{equation}\label{eq:3dpush}
    \mathcal{L}_{push}^{3D} = \frac{1}{C(C-1)}\sum_{c_A=1}^{C}\sum_{c_B=1,c_B \ne c_A}^{C} \left[ \delta_{push} - ||\mu_{c_A} - \mu_{c_B}||\right]^2_{+}
\end{equation}
\begin{equation}\label{eq:3dembed}
    \mathcal{L}_{3D}^{embed} = \lambda_{3D}^{pull}\mathcal{L}_{pull}^{3D} + \lambda_{3D}^{push}\mathcal{L}_{push}^{3D}
\end{equation}
\begin{equation}\label{eq:3doffset}
    \mathcal{L}_{offset}^{3D} = \sum_{i=1}^{s_1\times s_2} \mathbf{1}_{centerline}\left( \sigma(\Delta \hat{x}_i) - \Delta x_i \right)^2
\end{equation}
\begin{equation}\label{eq:3dheight}
    \mathcal{L}_{height}^{3D} = \sum_{i=1}^{s_1\times s_2} \mathbf{1}_{centerline}\left( \hat{h}_i - h_i \right)^2
\end{equation}

where $\Delta x_i$ and $h_i$ refers to the ground truth offset and height and $\Delta \hat{x}_i$ and $\hat{h}_i$ refers to the predicted offset and height and $\sigma$ refers to the sigmoid function. $\mathbf{1}_{centerline}$ is an indicator function that tells us whether a grid cell contains a centerline keypoint or not. $\mu_{c}$ denotes the cluster center for class $c$ in the embedding space. $C$ denotes the total number of classes. Therefore, the final 3D loss becomes:
\begin{equation}
    \mathcal{L}_{3D} = \mathcal{L}_{3D}^{embed} + \lambda_{3D}^{seg}\mathcal{L}_{3D}^{seg} + \lambda_{3D}^{offset}\mathcal{L}_{3D}^{offset} + \lambda_{3D}^{height}\mathcal{L}_{3D}^{height}
\end{equation}

\section{Results}
We evaluate the performance of two centerline detection baselines on the dataset created via the method proposed in \ref{sec:dataset}. We additionally examine the impact of the occlusion threshold $T_{occ}$ on the model outputs. Finally we conclude with findings related to the interpretability of attention models and also perform a qualitative analysis of the models on unseen data collected from a full-scale autonomous driving platform.

\textbf{Dataset:} The total split for the train, validation, and test sets corresponds to 659,720, 141,138, and 141,303 image frames, respectively. The labels for each frame include centerlines with both 2D pixel coordinates and corresponding 3D coordinates in the camera frame for the centerline keypoints. This dataset contains images from the Argoverse2~\cite{Argoverse2} which contains data from Austin, Detroit, Miami, Pittsburgh, Palo Alto, and Washington, D.C. We only use the front three cameras for our analysis as centerlines that are important for navigation are present in these three views but in practice all seven views can be used.

\textbf{Implementation:} We resize our images to the size of $576\times 1024$. There are a total of five labels generated from the above dataset for supervision. The 2D labels are generated via spline fitting to the keypoints on the image plane. The 3D labels are generated via spline fitting in 3D and then equidistant keypoints are selected with respect to the BEV grid centers. To reduce overlap in frames, we split the training set in windows of size 20 and sample a single frame randomly. In practice, we train on 32,986 frames per epoch for 70 epochs, effectively introducing variations in the training process while reducing training time. The learning rate used is $10^{-4}$ for the attention-based view transformation and $10^{-3}$ for the VRM.


\begin{figure}[h]
    \centering
    \begin{subfigure}{.5\linewidth}
      \centering
      \includegraphics[width=\linewidth]{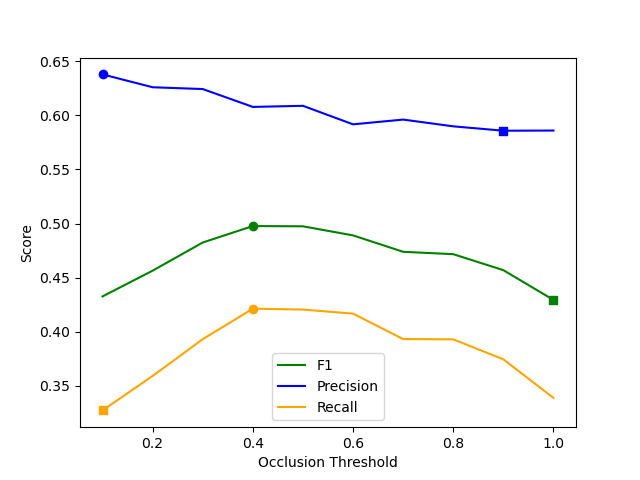}
      \caption{F1-score}
      \label{fig:voronoi-opath}
    \end{subfigure}%
    \begin{subfigure}{.5\linewidth}
      \centering
      \includegraphics[width=\linewidth]{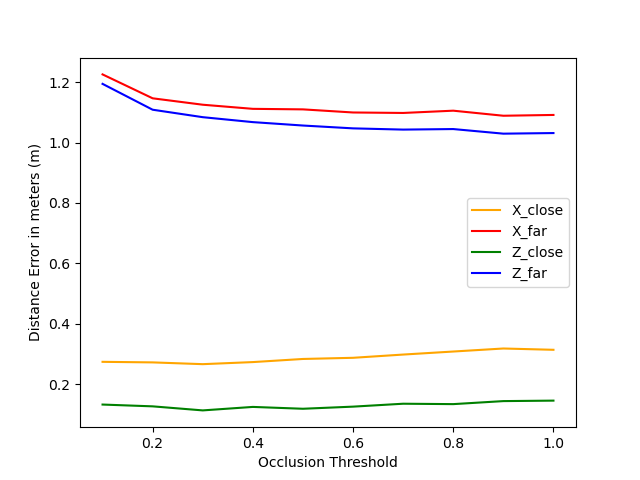}
      \caption{X/Z errors}
      \label{fig:voronoi-smooth}
    \end{subfigure}%
    \caption{Metrics for VRM with different $T_{occ}$}
    \label{fig:metric-tocc}
\end{figure}

\begin{figure*}[t]
  \vspace{0.2cm}
  \centering
   \includegraphics[width=.8\textwidth]{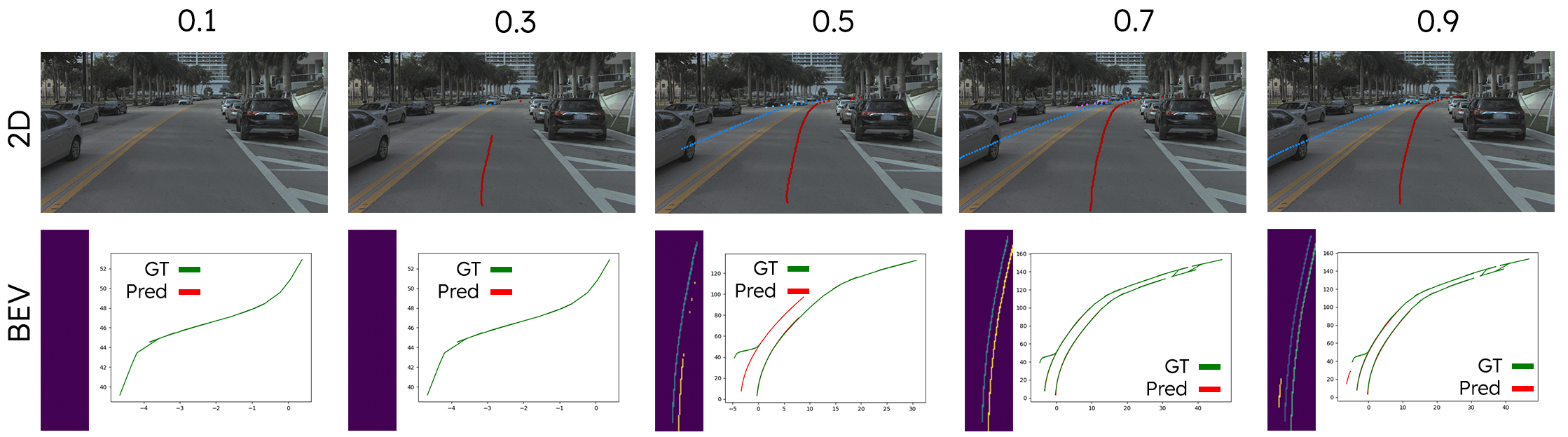}

   \caption{2D Image, BEV and 3D outputs of model trained with different $T_{occ}$ on Argoverse2 test set}
   \label{fig:tocc}
\end{figure*}

\begin{table*}[h]
\caption{VRM vs Attention on Argoverse2 test set}
\label{tab:att-bench}
\begin{center}
\begin{tabular}{|c||c|c|c|c|c|c|c|c|c|}
\hline
\textbf{Model} & \textbf{Inference Time} ($ms$) & $T_{occ}$ & \textbf{F1} & \textbf{Precision} & \textbf{Recall} & \textbf{X error} & \textbf{X error} & \textbf{Z error} & \textbf{Z error}\\
& & & & & & \textbf{near} & \textbf{far} & \textbf{near} & \textbf{far} \\
\hline
\hline
VRM & $11.819 \pm 0.501$ & $0.2$ & $0.456$ & $0.626$ & $0.359$ & $0.272$ & $1.147$ & $0.126$ & $1.109$\\
 & & $1.0$ & $0.429$ & $0.586$ & $0.339$ & $0.313$ & $1.092$ & $0.145$ & $1.032$\\
 \hline
 Attention & $19.076 \pm 1.184$ & $0.2$ & $0.486$ & $0.598$ & $0.410$ & $0.275$ & $1.155$ & $0.128$ & $1.117$\\
 & & $1.0$ & $0.439$ & $0.572$ & $0.357$ & $0.319$ & $1.099$ & $0.145$ & $1.036$\\
 \hline
\end{tabular}
\end{center}
\end{table*}

\subsection{Ablation: Occlusion Thresholds}
In Figure \ref{fig:metric-tocc} we compare the impact that various occlusion thresholds $T_{occ}$ have on the metrics using the VRM as the BEV feature converter. We use F1-Score and X/Z errors to benchmark our performance. The results indicate that quantitatively, the model with $T_{occ} = 0.4$ performs best. But qualitatively the results for higher thresholds achieve better performance in more occluded scenes. This is because smaller thresholds are stricter in the training process: occluded centerlines that can be recovered are often removed. As we can see in Figure \ref{fig:tocc}, models trained at higher thresholds perform better for the purposes of downstream navigation tasks by removing the erroneous centerlines seen in Figure \ref{fig:ucsd-viz} when no occlusion handling is present.

An important note to make is that as we increase the threshold, the benchmark becomes stricter and evaluates a higher number of occluded centerlines. While we observe a decrease in F1 score due to highly occluded scenarios, the quality of the predictions at higher thresholds is consistent and robust to moderate and even certain severe occlusions. The first row in Figure \ref{fig:tocc} shows the 2D image detections of the centerline keypoints. The second row in Figure \ref{fig:tocc} shows the BEV and 3D detections of the centerlines along with ground truth.

\subsection{Models and Interpretability}
\begin{figure}[h]
  \centering
   \includegraphics[width=.7\linewidth]{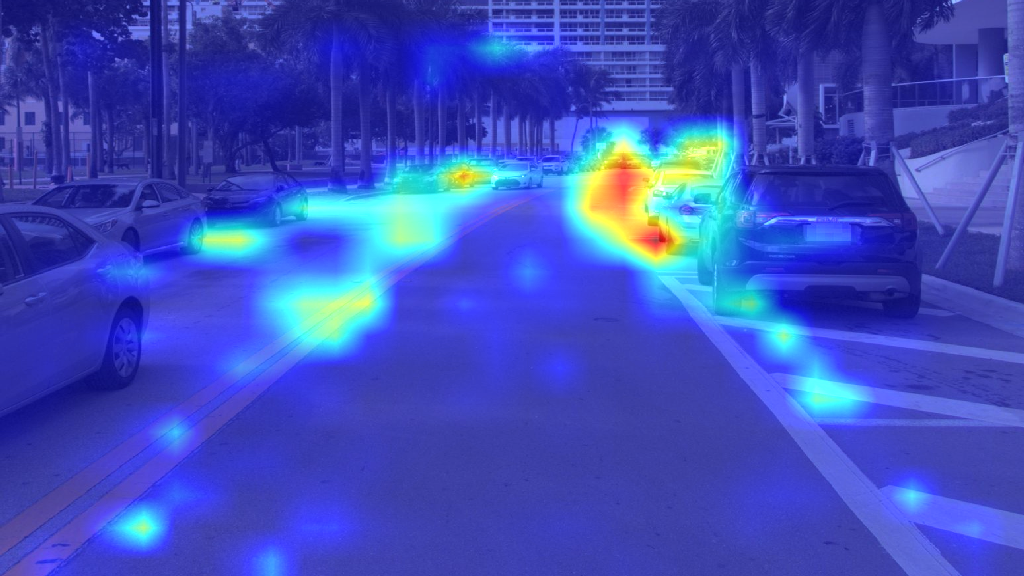}

   \caption{Attention weights for Argoverse2 test set}
   \label{fig:att-vis}
\end{figure}

We experiment with replacing the VRM with a Multi-Head Attention(MHA) based BEV conversion, similar to the vision transformer. We achieve competitive results with both models with one noticeable difference as seen in Table \ref{tab:att-bench}. The attention-based models generally achieve better recall but worse precision when compared to the VRM over all $T_{occ}$. One advantage of the attention-based model is that we can visualize the attention maps and gain interpretability into which features matter for centerline detection and how the occlusion thresholds affect this. As seen in Figure \ref{fig:att-vis} the model seems to focus more on the  lanes and the road edges rather than the center itself, which is noteworthy as this is consistent with how human drivers try to determine the center of the road.



\subsection{UCSD experiments}
We also test the network models using a full-scale autonomous driving platform at the UC San Diego campus to check its generalization to different camera configurations. One big motivation for choosing a network that takes a single camera image as an input rather than all views is that most networks overfit to specific sensor configurations, leading to worse results on out-of-distribution data if all views are fed simultaneously to the network. The baseline network generalizes well in terms of the 2D image predictions, as can be seen in Figure \ref{fig:ucsd-viz}. However, the network experiences challenges for 3D predictions as expected due to changes in the camera location. Nevertheless, 3D points can be inferred via late-fusion methods across multi-view camera perspectives.



Both models implemented in this work are additionally evaluated in terms of real-time capabilities. The inference times reported in Table \ref{tab:att-bench} meet real-time constraints for driving applications. The VRM provides advantages regarding inference time; however, the attention-based approach can boost performance by reducing false negatives. The inference time experiments are performed using an RTX 4090 GPU.

\section{CONCLUSIONS}
In conclusion, this research sets the stage and represents a significant step toward enhancing road topology estimation in dynamic urban driving scenarios. Creating a substantial and publicly available dataset using automatic label generation and occlusion handling strategies provides valuable resources for further advancements in this field. The findings offer insights into the challenges associated with occlusion, with demonstrated practicality and adaptability of these methods. For future work, integrating real-time centerline detection models with nominal   maps could offer a promising avenue for enhancing the accuracy and reliability of road topology determination, making it even more valuable for autonomous driving and urban planning applications. 




\section*{ACKNOWLEDGMENT}
We sincerely appreciate the support from Nissan Motor Co. We also acknowledge the input and feedback from various members of the Autonomous Vehicle Laboratory at UC San Diego, including Henry Zhang and Seth Farrell.




{\small
\bibliographystyle{./IEEEtran}
\bibliography{IEEEcitation}
}

\end{document}